# OGInfra: Geolocating Oil & Gas Infrastructure using Remote Sensing based Active Fire Data

**Samyak Prajapati**
*Department of Computer Science & Engineering*
UT Arlington, USA
sxp5756@mavs.uta.edu

**Amrit Raj**
*Department of Computer Science & Engineering*
NIT Delhi, India
181210008@nitdelhi.ac.in

**Yash Chaudhari**
*Department of Computer Science*
University of Central Florida, USA
yesh0607@knights.ucf.edu

**Akhilesh Nandwal**
*Department of Computer Science & Engineering*
NIT Delhi, India
181210006@nitdelhi.ac.in

**Japman Singh Monga**
*Department of Computer Science & Engineering*
NIT Delhi, India
181210024@nitdelhi.ac.in

**Abstract**

Remote sensing has become a crucial part of our daily lives, whether it be from triangulating our location using GPS or providing us with a weather forecast. It has multiple applications in domains such as military, socio-economical, commercial, and even in supporting humanitarian efforts. This work proposes a novel technique for the automated geo-location of Oil & Gas infrastructure with the use of Active Fire Data from the NASA FIRMS data repository & Deep Learning techniques; achieving a top accuracy of 90.68% with the use of ResNet101.

## INTRODUCTION

In today's world, where data is king, approaches that incorporate data-driven analytics can generate insights that might just not be possible with basic intuition; this can be attributed to the inability of comprehending data of such large scale, or due to the fact that the monotonicity of manual analysis of data is an unnerving task. The combination of deep learning approaches with such menial tasks grants us the ability to create applications that can mimic a fraction of human intuition by learning the important features present in the data.

Remote sensing, since its conception was a method to detect and track problems that plague us, humans, as a whole. While it may not be possible to solve these problems directly, remote sensing has the incredible potential to generate a prognosis of the "diseases" affecting the Earth. This technology has gained traction as a powerful "macro-level" monitoring tool in several domains such as precision agriculture, forestry research, public policy management, climate change, and much more [1][2][3][4]. This approach also has implications in the Oil & Gas sector, where it can be used to track methane emissions as exhibited in the works of Schneising et al. [5]

The NASA FIRMS dataset is a collection of infrared "hotspots" that are considered to be active fire zones; this is accomplished by collecting data from sensors aboard multiple satellites

(MODIS aboard the NASA Terra & Aqua, and VIIRS aboard the Suomi-NPP & NOAA 20). The combination of these sensors and indirectly, this dataset, allows for the "Near Real Time" monitoring of active fires. The existence of such data, targeting thermal hotspots has multiple implications in several unrelated fields, one of them is where the FIRMS data is used to generate insights on active fires and is actively being used to locate and target active wildfires in the US and Canada [6].

There exists another such specialized dataset, known as the Global Oil and Gas Infrastructure (GOGI) geodatabase, which was developed in collaboration with the National Energy Technology Laboratory (NETL) and the Environmental Defense Fund (EDF) with the goal of creating an open data foundation to tackle a range of stakeholder questions and needs [7]. However, there might arise certain geo-political situations that force the development of guerilla infrastructure of the Oil & Gas sector in a rapid manner; an example being the recent Russian invasion of Ukraine, where essential fuel depots and pipelines would have to be placed in order to support the ground forces. In such cases, remote sensing can play a vital role in tracking the developments of such infrastructure, and thus, by extension, it can help us track the events of a situation as they unfold.

Infrastructure that supports this industry often has multiple flare stacks to burn off excess flammable gas and noxious fumes, therefore, it is often used as a safety device. This work relies on the presence of similar structures and aims towards developing a pipeline for the automated detection of Oil & Gas Infrastructure using remote sensing and deep learning techniques; a workflow summary is presented in Figure 1.

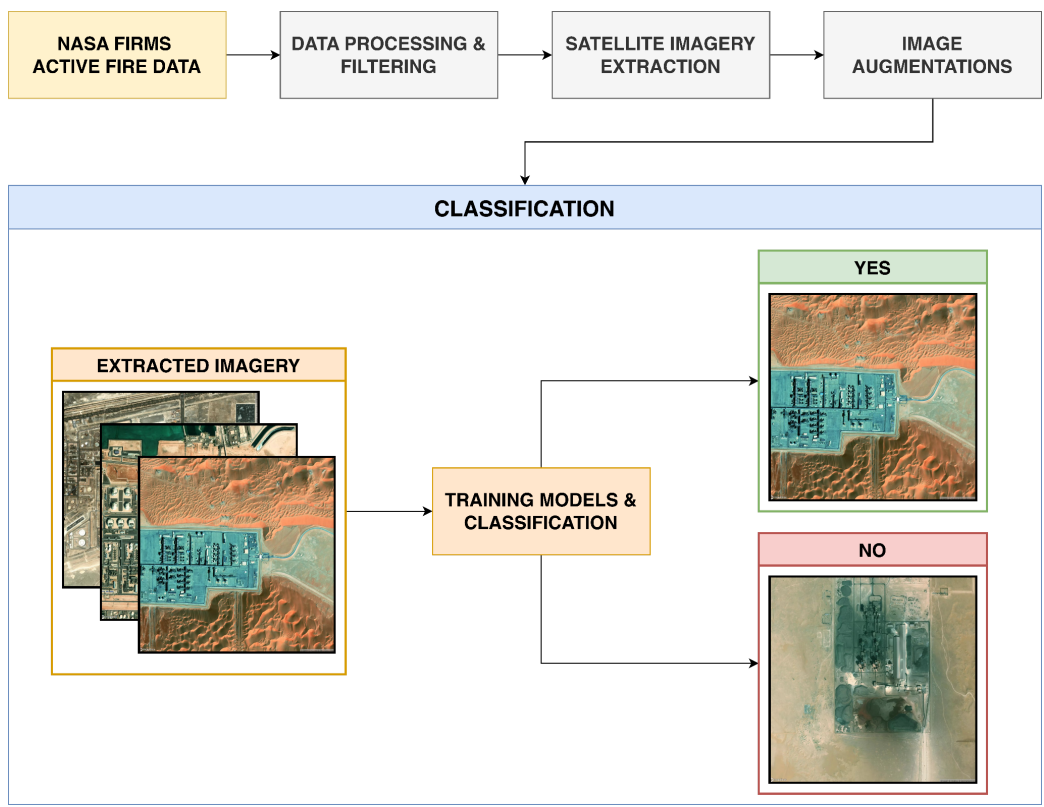

*Figure 1: Workflow Summary*

# RELATED WORKS

While there has been limited work done in this domain, there still are certain works, such as that of Ben Lord [8] where they vocated for the use of hyperspectral sensors for remotely sensing onshore oil and gas reserves. They utilized spectral analysis which involves the analysis of energy absorbed and reflected by hydrocarbon systems on the earth's surface. The dataset used in that study was collected from the United States Geological Survey (USGS), SPECTIR, and ITRES. This data was then modified to correct geometric and radiometric abnormalities and then atmospheric algorithms were implemented through Harris Geospatial's ENVI program to extrapolate and smoothen the data. The classification algorithms were then amalgamated into a decision tree, creating a final classification result. The classification algorithms included spectral angle mapper (SAM), linear spectral unmixing (LSU), continuum removal, and band ratios.

Lu Shen et al. [9] tried to assess anthropogenic methane emissions using TROPOMI (Tropospheric Monitoring Instrument) data on a national and regional basis. They also focussed on different sectors that account for methane emissions such as oil/gas, coal, landfills, etc. Their work mainly revolved around the area of eastern Mexico. One of the major issues suggested by this study was uncertainty in the methane emission estimates based on the bottom-up inventory. Hence, to establish and counter this issue, in this study, the methane emissions were estimated by the process of atmospheric inverse analysis by using the observations obtained in column averaged methane mixing ratio from TROPOMI. The results of this study suggest that unaccounted emissions of southern onshore basins are one of the major factors which result in unreliability in estimates of bottom-up inventory. The study called for stronger methods to reduce methane emissions owing to the large discrepancies found in oil/gas related methane emissions in the country.

Hao Sheng et al. [10] tried to estimate and quantify the methane emissions from the Oil Refineries making use of data obtained from satellites, namely SCIAMACHY, GOSAT, and TROPOMI. Authors in this research paper claimed methane emissions from Oil and Gas (O&G) industries to be responsible for global warming by up to a quarter extent. They mainly focused on locating Oil and Gas Refineries that were active and this research mainly emphasized the United States. They focused mainly to locate Oil Refineries in remotely sensed images as they have a dominant footprint and consistent features. Their work mainly concentrated on Oil and Gas infrastructures located in the United States. For this, they made use of 149 locations of oil refineries obtained from the Enverus Drillinginfo database. The dataset was predominantly biased toward the positive sample of aerial images of oil refineries, therefore additional negative samples were added to the dataset, keeping in mind they do not resemble oil refineries or the landscapes found near the oil refineries. Their research demonstrated and paved the way to fill the gaps between real data and the data present in the famous four public datasets of Oil and Gas infrastructure using deep learning techniques.

# METHODOLOGY

Through this study, we propose a novel technique that can be used to detect oil and gas infrastructure in certain landscapes using the combination of remote sensing and deep learning. This proposed methodology is a three-fold approach, namely, (i) Data Collection, (ii) Data Processing, and (iii) Model Training.

1. **Data Collection**

    Data extraction was performed from the NASA FIRMS (Fire Information for Resource Management System) data repository [11][12]. This repository contains data from the VIIRS sensor and the MODIS sensor. The VIIRS (Visible Infrared Imaging Radiometer Suite) sensor data was first extracted from the FIRMS repository for the entirety of 2021. The VIIRS sensor can collect data from a total of 22 spectral bands between 0.412 μm and 12.01μm, with 16 moderate resolution bands (M-bands) with a spatial resolution of 750m; 5 high resolution imaging resolution bands (I-bands) with a spatial resolution of 375m at nadir and a single day-night band (DNB) with a nominal spatial resolution of 750m. [13] The data used in this work was collected majorly from two satellites, the Suomi National Polar-orbiting Partnership (Suomi NPP) and the NOAA-20 weather satellites from 2021-01-01 to 2021-12-31 (One Calendar Year); certain limitations were also put on the extracted data to ensure the quality of data being used, this will be further discussed in the subsection titled "Data Processing".

    The static satellite imagery for each relevant coordinate was fetched from the MapBox API using the *satellite-v9* style. Each image had a zoom level of 15 and was saved as a PNG with a resolution of 1000x1000 pixels.

2. **Data Processing**

    This section describes the data processing pipeline used on cartesian coordinate data from the FIRMS repository and the satellite imagery data. The VIIRS 375m data distributed by NASA FIRMS contains many attributes but certain ones, such as *latitude*, *longitude*, *brightness*, *confidence*, *and bright_t31* are of importance for this work [14]. Certain thresholds for these features were chosen empirically to satisfy the base criterion of having enough data points to achieve trainable data for the model; due to this, the following constraints were put on the data,

    - *brightness >= 300,*
    - *bright_t31 >= 270,*
    - *confidence == 'h' or confidence == 'n'*

    Keeping these constraints in mind, a large number of coordinates were fetched from the combined database of SUOMI-NPP and NOAA-20 satellites since the temporal span of the data was one entire year. To handle the vastness of this data and to keep computational resources in check, the data from a single satellite was imported, processed with constraints, and then repeated for the other satellite; this is summarized in Figure 2.

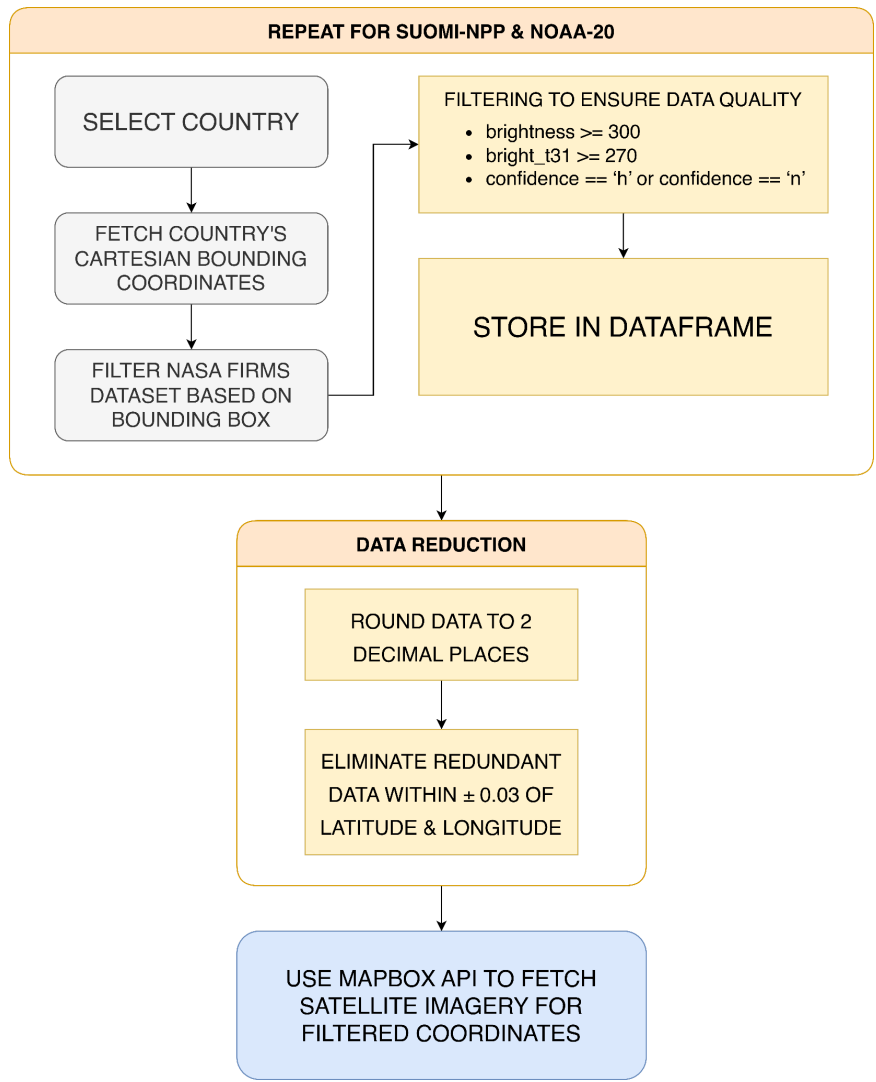

*Figure 2: Data Processing Pipeline*

3. **Model Training**

In order to empirically evaluate the effectiveness of the proposed methodology, four models of varying features (depth, presence of skip connections) were trained and compared based on Top-1 Accuracy, AUC, and F1 Score. In order to do so, two classes of images were selected, where the "positive" image class had the presence of Oil & Gas infrastructure and the "negative" image class contained images samples that did not contain any relevant infrastructure. As discussed previously, the works of Sheng et al. [10] were also appended to our custom dataset to boost the image count, however, this led to a gross imbalance between the classes. In order to ameliorate this imbalance, the positive images collected from the NASA FIRMS repository were augmented with vertical & horizontal flipping and rotations of 90, 180, and 270 degrees, and this helped increase the positive image count from 274 to 1644; thus reducing the level of class imbalance.

Table 1 below expands on the model characteristics of the chosen models and delineates the reasoning behind choosing the specified model. The Relative Inference Time is calculated by dividing the Inference Time per Image (on a batch size of 16) from the works of Bianco et al. [15] by the number of CUDA cores of the GPU used in assessing the performance (NVIDIA Titan X Pascal GPU with 3840 CUDA cores).

*Table 1: Model Parameters*

| Model | Parameters | Depth | Relative Inference Time (GPU) |
|---|---|---|---|
| **AlexNet** | 62M | 8 | 0.047 ms |
| **VGG-16** | 138.4M | 16 | 0.573 ms |
| **ResNet-101** | 44.7M | 209 | 0.630 ms |
| **InceptionResNetV2** | 55.9M | 449 | 1.352 ms |

We wanted to assess the performance of this methodology on models of varying depth and complexities, thus our chosen models range from AlexNet to InceptionResNetV2, with varying depth and inference times.

In our work, these models were trained on an NVIDIA Tesla T4 with 2560 CUDA cores with the Keras library; Early Stopping was also implemented on validation loss with a minimum delta of 0.002 and patience of 5 epochs.

## RESULTS

The processed data was used to train the four chosen CNN-based models, AlexNet, VGG-16, ResNet-101, and InceptionResNetV2. Performance metrics such as Binary Cross-Entropy (Loss), Top-1 Accuracy, Area under the ROC curve, and F1 scores were the metrics chosen to reflect the performance of the models on the classification task; results for each model are tabulated in Table 2.

*Table 2: Performance Metrics of Models*

| Model | Loss | Accuracy | AUROC | F1 Score |
|---|---|---|---|---|
| **AlexNet** | 0.6324 | 0.7280 | 0.7225 | 0.7283 |
| **VGG-16** | 0.2474 | 0.8967 | 0.9627 | 0.8975 |
| **ResNet-101** | 0.2461 | 0.9068 | 0.9508 | 0.9063 |
| **InceptionResNetV2** | 0.2945 | 0.8791 | 0.9491 | 0.8800 |

## DISCUSSIONS

We can see that AlexNet has the lowest performance amongst the four models, and this is obviously due to the simplicity of the model. We must note that "simple feed-forward" models like the VGG-16 offer comparable performance to the complex models, such as ResNet-101 and InceptionResNetV2, both of which feature skip connections, which allow the model to learn efficiently. Unlike the expectation, that a model with significantly higher depth would yield impeccable performance towards the task; however, the contrary is observed in our results. ResNet101, with almost half of the depth of InceptionResnetV2, performs better with a testing accuracy of 90.68% and an F1-Score of 0.9063. It may be attributed to the fact the elevated complexity of the InceptionResNetV2 model requires a significantly higher level of fine-tuning, and in cases like these, where we train for larger epochs, and with reduced learning rates, we risk overfitting our model.

The aim of this work was to act as a proof-of-concept, and not to achieve the highest possible performance metrics, therefore, it can be expected that there are other models, with varying parameters, that can perform better than those selected in this work, and this is a subject for future research in this domain.

It is vital that we note that this approach relies on the presence of detectable flames from the flaring stack present onsite. Since this work makes use of Active Fire Data from the data repository, if there is no physical flame present, the site can not be located. This is a recognized shortcoming of this approach; however, one way to circumvent this is to collect data over multiple days. The NOAA-20 operates almost 50 minutes ahead of the Suomi NPP, and their combination allows for full global coverage every 12 hours, therefore it is possible to collect real-time data and process it to attain imagery with a certain lag. This can help collate data and thus, the usage of this methodology would be possible.

## CONCLUSIONS

A lot of progress has been made in the field of remote sensing and developing new sensors that can be used to capture a multitude of data. This paper makes use of one such sensor (VIIRS) that can be used to detect active fire data using infrared sensors. We propose a novel approach that is geared toward geolocating Oil & Gas infrastructure by using remote sensing data coupled with a deep learning methodology. Our implementations achieved top accuracy of 90.68% when training and testing on limited models. While this approach was tailored towards the arid desert landscape of the Middle East countries due to computing limitations, it can easily be expanded to focus on other parts of the world, each with varying landscapes. This methodology can be utilized to track the development of infrastructure dedicated to the Oil & Gas industry through a passive approach that does not require any manual intervention. Sufficient levels of automation can perhaps be helpful in understanding and monitoring the guerrilla expansion of a nation into the energy sector and help policymakers to draft and implement effective laws.